\documentclass[
]{ceurart}

\usepackage{caption}
\usepackage{subcaption}
\usepackage[htt]{hyphenat}

\usepackage{booktabs}
\usepackage{array}
\usepackage{makecell}

\usepackage{listings}
\begin{document}

\copyrightyear{2024}
\copyrightclause{Copyright for this paper by its authors.
  Use permitted under Creative Commons License Attribution 4.0
  International (CC BY 4.0).}

\conference{CHR 2024: Computational Humanities Research Conference, December 4–6, 2024, Aarhus, Denmark}

\title{Once More, With Feeling: Measuring Emotion of Acting Performances in Contemporary American Film}

\author[]{Naitian Zhou}[%
orcid=0009-0005-1991-2258,
email=naitian@berkeley.edu,
]
\cormark[1]
\address[]{School of Information, UC Berkeley, USA}

\author[]{David Bamman}[%
orcid=,
email=dbamman@berkeley.edu,
url=,
]
\cortext[1]{Corresponding author.}

\begin{abstract}
Narrative film is a composition of writing, cinematography, editing, and performance. While much computational work has focused on the writing or visual style in film, we conduct in this paper a computational exploration of acting performance. Applying speech emotion recognition models and a variationist sociolinguistic analytical framework to a corpus of popular, contemporary American film, we find narrative structure, diachronic shifts, and genre- and dialogue-based constraints located in spoken performances.
\end{abstract}


\maketitle

\section{Introduction}

Film is rich in its supply of semiotic resources, communicating meaning from the interaction of language (encoded in a script), visuals (choices of composition, blocking, cinematography), sound and more.  Much computational work has arisen to examine slices of this semiotic field, including measuring how gender stereotypes or plot arcs are reflected in dialogue \cite{yuUnpackingGenderStereotypes2022a,schmidtPlotArceologyVectorspace2015,kayhaniMoviesEmotionalAnalysis2020} or how visual features like color variance and shot length constitute genre \cite{rasheedUseComputableFeatures2005,guhaComputationallyDeconstructingMovie2015}. One critical area, however, that has been neglected in this study is the role of \emph{performance} in creating meaning.

As \citet{naremoreActingCinema1988} notes, film is a medium in which meaning is \textit{acted out};
an acting performance provides a semiotic frame through which we can understand the events that unfold. Given the fixed text of a script, the rendering of the final performance is an interpretive process in which the actor, director and editor jointly imbue the words with additional meaning.  In this view, the {same} line of dialogue exhibits variation in meaning when performed  in distinct diegetic contexts. As one example, consider the following line in \textit{Knives Out} (2019):

\begin{center}
    ``I'm warning you.''
\end{center}

Much of the film revolves around these three words, overheard in a conversation between the wealthy Harlan Thrombey and his grandson, Ransom. The line is uttered by multiple characters as the film unfolds: angrily shouted by Ransom, somberly recalled by the eavesdropper, and gleefully recounted by inspector Benoit Blanc upon solving the crime. Even a single line, located within a single diegetic event, has great capacity for meaning-making in performance.

When viewed in this light, we can apply the analytical framework from variationist sociolinguistics to better understand this space of performance. Given a fixed line of dialogue (equivalent to a linguistic variable), a performance entails a choice --- a selection from the set of possible variants. It is this choice, and the meaning contained within, which we study.

In this work, we design computational models to explore this form of variation by considering the emotional range of performances in contemporary American film, exploring in particular the tension between \textit{what} characters say and \textit{how} they say it. As distinct from prior work in the computational humanities that has measured emotion from text alone \cite{kayhaniMoviesEmotionalAnalysis2020,kimInvestigatingRelationshipLiterary2017}, we measure acted emotion from speech, allowing us to disentangle the emotion present in the script from the choices made in creating the performance. 

Using a speech emotion recognition model, we construct a parallel dataset of spoken performances (\textit{utterances}) aligned with the text of the words being spoken (\textit{dialogue phrases}). This dataset allows us to isolate and examine how performances vary in meaning from their \textit{para}linguistic features in addition to the textual meaning of the screenplay.  We use this dataset to carry out several case studies exploring variation in performance in American film:

\begin{enumerate}
    \item First, we carry out a structural analysis of emotion as performed over narrative time. Doing so allows us to characterize film as \textit{performance text}, relating emotional performance to larger narrative structure.
    \item Second, we study diachronic variation by comparing emotionality of films across release years, testing the degree to which performances have intensified over time (following Bordwell's theories of visual style\ \citep{bordwellIntensifiedContinuityVisual2002}).
    \item Finally, we examine the \textit{capacity} for performance by constructing a novel measure of \textit{emotional range} for an utterance---the space of possible emotions that can be performed. In doing so, we demonstrate how both contextual (genre) and textual (dialogue) aspects of film can carry constraints and affordances on acting performance.
\end{enumerate}

In this work, we use computational methods to survey how both textual and contextual variables inform and reflect the performances rendered on screen.

\section{Methods}
\label{sec:methods}

In order to perform our analysis, we need to construct an aligned dataset of actor performances (\textit{utterances}), the text of the words they speak (\textit{phrases}), and the emotions in each utterance. We create a pipeline that takes as input a set of full-length movies, and outputs time-aligned transcriptions for utterances, their emotion labels, and groups of semantically similar phrases.

\subsection{Preprocessing pipeline}

We first construct a data pipeline to segment and transcribe utterances from movie dialogues. The pipeline takes as input a set of MP4 files, where each file is one digitized film. Our analysis takes place in the speech and text modalities, so we use \texttt{ffmpeg} to extract the audio track.

We use the pyannote\footnote{\url{https://huggingface.co/pyannote/speaker-segmentation}} segmentation model to detect continuous, single-speaker speech segments. Then, we use faster-whisper\footnote{\url{https://github.com/SYSTRAN/faster-whisper}} to transcribe each speech segment. 
Because pyannote speech segments are based on voice activity detection and silences, it can label extended, multi-sentence speech as a single segment. For our analysis, however, we are interested in utterances as a discursive unit. If a character makes an assessment, then poses a question, we would like to split these into two distinct utterances. As a middle ground between raw voice activity detection and segmenting discursive units, which requires complex conversational understanding, we perform a post-processing step where we further split speech segments by sentence boundaries derived from the transcriptions. Because whisper is an end-to-end model that does not produce fine-grained time alignments, we then use a speech-to-text fine-tuned wav2vec2\footnote{\url{https://huggingface.co/facebook/wav2vec2-base-960h}} to perform word-level time alignment between the transcription and the audio, then split the audio based on sentence boundaries generated by \texttt{syntok},\footnote{\url{https://github.com/fnl/syntok}} a fast, rule-based sentence segmenter.

To prevent the end credit sequences from interfering with the results, we detect when the end credits begin by performing optical character recognition (OCR) on the shots in a movie and identifying long continuous sequences of shots that contain large amounts of text. We trim the movie to the beginning of the end credits.

\subsection{Speech emotion recognition}

To perform speech emotion recognition, we use a wav2vec2 large model without any task-specific fine-tuning to extract audio features. Then, we train a classification head to perform seven-way emotion classification, based on the Ekman model \cite{ekmanArgumentBasicEmotions1992} of six basic emotions (anger, disgust, happiness, sadness, fear, and surprise) and a neutral label. To train these models, we use the MELD dataset, which contains 1,000 sampled dialogues from the TV series \textit{Friends} \cite{poriaMELDMultimodalMultiParty2019}.

We experiment with two classification settings: an \textit{utterance-level} model which makes predictions based on only the speech features of the input utterance and a \textit{conversation-level} model which includes the speech features of both the input utterance and its surrounding utterances.

In both cases, we use a pretrained wav2vec2 model as a backbone model for generating vector representations of each utterance. Because wav2vec2 creates an embedding for each audio frame (roughly 20ms of speech), we follow prior work in computing utterance embeddings by averaging across all timestamps within an utterance \cite{pepinoEmotionRecognitionSpeech2021}. We compute embeddings from the attention activations at each layer of the wav2vec2 model instead of just taking the last-layer activations; prior work has shown that, for paralinguistic tasks such as emotion recognition, early- and intermediate-layer activations are more useful than later layers \cite{pepinoEmotionRecognitionSpeech2021,shorTRILLssonDistilledUniversal2022}. At the end of the embedding step, each utterance is represented by a set of 25 768-dimensional vectors.

\subsubsection{Utterance-level emotion recognition}

We implement the utterance-level model from \citet{pepinoEmotionRecognitionSpeech2021} and match the reported performance. We first take a weighted average of layer activations for a given utterance; these weights are learned during training. Then, we apply a fully-connected classification head to produce a probability distribution over the seven emotion labels. Unlike the original paper, we do not use features from the initial convolutional layer of the pre-trained model; we use only the attention head activations.

\subsubsection{Contextual emotion recognition}

We also train an contextual model which uses a bidirectional LSTM to predict the emotion of utterances within the context of a conversation. To do so, we define conversations as groups of utterances where each occurs within 3 seconds of the next. In the MELD dataset, there are 1,478 conversations in the training split according to this criterion.

For each conversation, we predict the emotion labels of all utterances in the conversation by first passing  weighted activations through the biLSTM before applying the classification head to each hidden state. As before, the weights of activations are learned during training.

\subsubsection{Evaluation}

We expect the movie data to be similar in nature to the MELD dataset, since both consist of professionally produced and acted clips. However, to ensure that our models do not experience domain shift despite the greater range in release year and setting of the film corpus, we evaluate these models on the test split of the MELD dataset as well as a manually collected dataset consisting of 333 clips from a subset of 35 contemporary American films. Each clip was a conversation with at least 2 utterances, where conversations were identified with the same heuristic used to construct training data for the contextual model. This resulted in a final evaluation dataset of 2,157 utterances with emotion labels. The clips were labeled by two annotators: 51 clips were labeled by both annotators and 250 clips were labeled by a single annotator. The Krippendorff's $\alpha$ between the two annotators was 0.334, and the Fleiss' $\kappa$ was 0.333, which matches the agreement of the MELD dataset.

Table~\ref{tab:model_comp} shows the evaluation results on the MELD and Movies datasets. The models perform comparably to each other, and comparably across evaluation datasets. This performance also approaches the state of the art on MELD for audio-only models. Because the performance of the contextual model is slightly higher, we use its inference outputs for our analysis.

\begin{table}[h]
    \centering
  \renewcommand{\arraystretch}{1.2} %
    \caption{Model comparison for MELD test dataset and our movies dataset, along with 95\% bootstrap confidence interval bounds.}

    \begin{tabular}{c|cc|ccc}
        \toprule
        & \multicolumn{2}{c|}{\textbf{MELD}} & \multicolumn{2}{c}{\textbf{Movies}} \\
        & Accuracy & Weighted F1 & Accuracy & Weighted F1 \\
        \midrule
        Utterance & 0.476 [0.451, 0.501] & 0.455 [0.432, 0.478] & 0.449 [0.429, 0.469] & 0.434 [0.413, 0.456] \\
        Contextual & 0.494 [0.472, 0.516] & 0.456 [0.441, 0.472] & 0.488 [0.468, 0.508] & 0.450 [0.428, 0.472] \\
        \bottomrule
    \end{tabular}
    \label{tab:model_comp}
\end{table}

\subsection{Identifying dialogue phrase groups}

One powerful aspect of this dataset is that we align actor performances to the words that they speak. To account for variation in how highly semantically similar phrases can be realized, we cluster together phrases with high semantic similarity. We use the \texttt{sentence-transformers} library to compute sentence embeddings of utterances and cluster them with the Leiden community detection algorithm \cite{traagLouvainLeidenGuaranteeing2019}. Table~\ref{tab:phrase_groups} shows some examples of phrases that are grouped together. We expect the phrases in each group to have similar prior distributions of emotion.

\begin{table}[h]
    \centering
      \renewcommand{\arraystretch}{1.2} %
    \caption{Examples of utterances which are clustered into dialogue phrase groups.}

    \begin{tabular}{p{0.8\linewidth}}
        \toprule
        \textbf{Phrase Groups} \\ 
        \midrule
        ``Let's go, let's go, let's go!'', ``Let's go, let's go!'', ``Let's go right now go go'', ``Go, let's go, let's go.'', ``Okay guys, let's go.'' \\ 
        \midrule
        ``Oh, pleasure to meet you.'', ``It's so nice to finally meet you.'', ``It is a pleasure to finally meet you.'', ``Oh, it's nice to meet you.'', ``It's so nice to meet you!'' \\ 
        \bottomrule
    \end{tabular}
    \label{tab:phrase_groups}
\end{table}

\section{Analysis}

\label{sec:data}

The above pipeline measures the emotions performed in an utterance and ties each utterance to the text being spoken. We apply this methodology to a large corpus of contemporary, popular American films \cite{bammanDiversityHollywood} in order to study the variation of emotion within and between them.

Our corpus consists of the top-50 live-action, narrative films by U.S. box office from 1980-2022. We supplement these with films nominated for ``Best Picture''-equivalent awards by one of six organizations in those years: Academy Awards, Golden Globes, British Academy of Film and Television Arts, Los Angeles Film Critics Association, National Board of Review, and National Society of Film Critics. We only include English-language films in this analysis, resulting in a total of 2,283 films.

\subsection{Film as performance text}

\citet{plantingaMovingViewersAmerican2009} describes how emotionality can reflect narrative structure --- emotionally-charged events can serve as catalyst to disrupt the expository ``stable state'' and set the narrative in motion. Much attention has been paid to characterizing narratives in literature and film in terms of emotionality using trajectories of sentiment \cite{jockersSyuzhetExtractSentiment2015} and emotion \cite{reaganEmotionalArcsStories2016,kayhaniMoviesEmotionalAnalysis2020,hipsonEmotionDynamicsMovie2021,vishnubhotlaEmotionDynamicsLiterary2024}; these have focused on the  emotion encoded in text. Audiences of movies, however, are not directly exposed to that text; their experience is mediated by the performance. In order to study this more directly, we turn our attention to characterizing narratives with emotion as performed.

We study the distribution of emotions in utterances over the course of a movie. How do the prevalence of emotions shift over narrative time? Similar to previous work on dialogue in screenplays, we ask if there are emotional regularities across films \cite{hipsonEmotionDynamicsMovie2021}. We examine first the \textit{emotionality} of utterances ---  the average probability that an utterance is not neutral --- before looking more closely at how specific emotions are distributed temporally. We plot the average probability of an emotion label for an utterance in intervals of 5 percent, expressed as a percentage of the full run-time of the film. Specific emotions are measured as proportions of the emotional labels, excluding the neutral label.

\begin{figure}
    \centering
    \begin{subfigure}[b]{0.45\textwidth}
        \includegraphics[width=\textwidth]{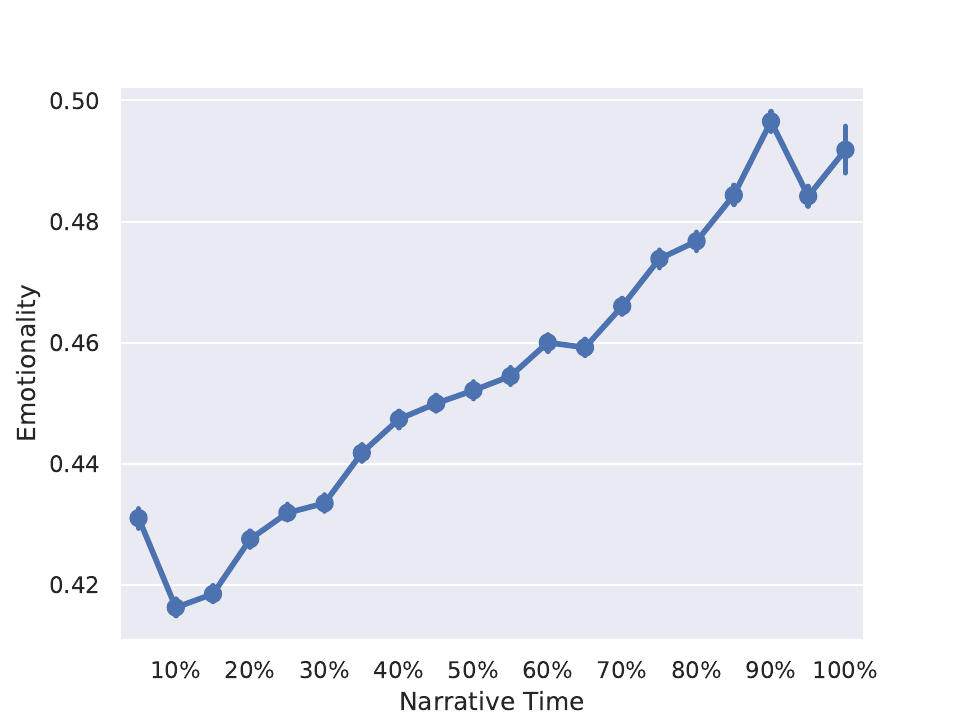}
        \caption{Emotionality}
        \label{fig:neutral_decline}        
    \end{subfigure}
    \begin{subfigure}[b]{0.45\textwidth}
        \includegraphics[width=\textwidth]{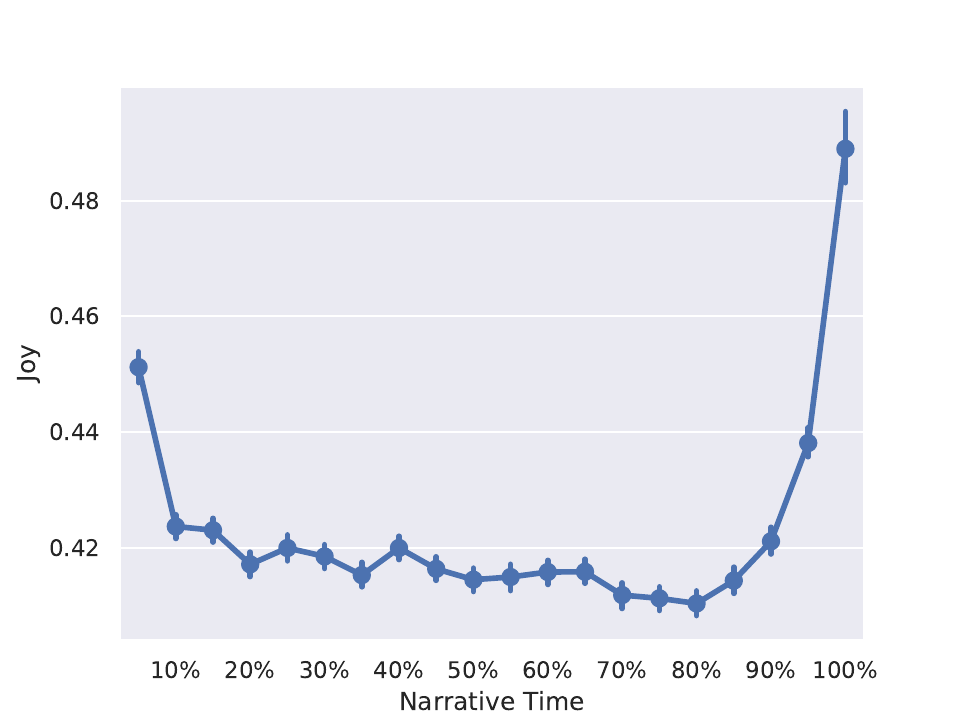}
        \caption{Joy}
        \label{fig:joy_narr}        
    \end{subfigure}
    \begin{subfigure}[b]{0.45\textwidth}
        \centering
        \includegraphics[width=\textwidth]{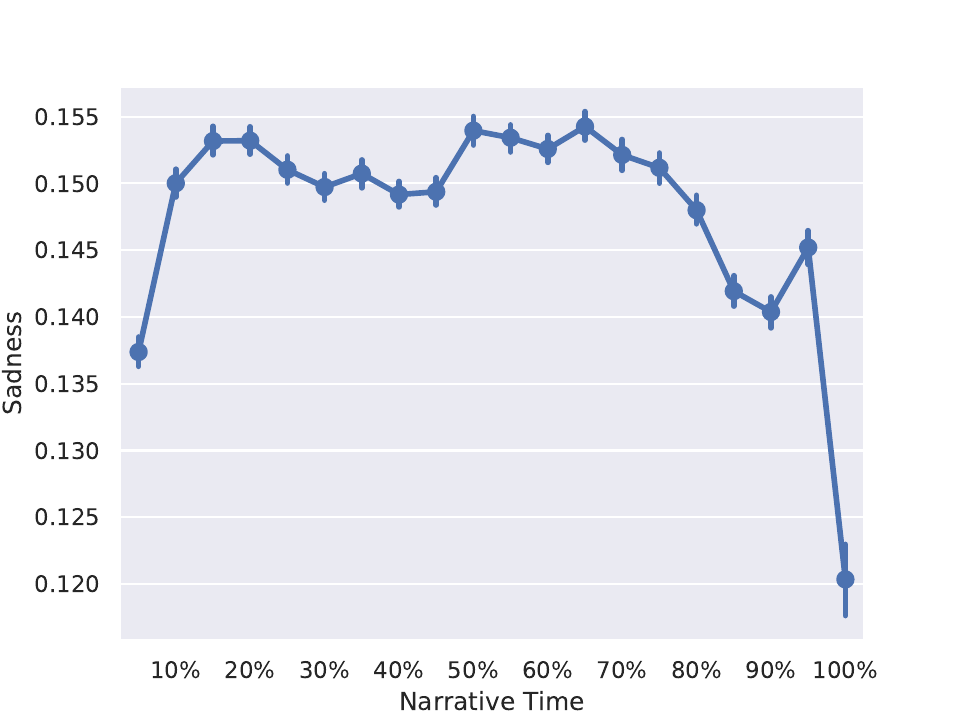}
        \caption{Sadness}
        \label{fig:sadness_narr}
    \end{subfigure}
    \begin{subfigure}[b]{0.45\textwidth}
        \includegraphics[width=\textwidth]{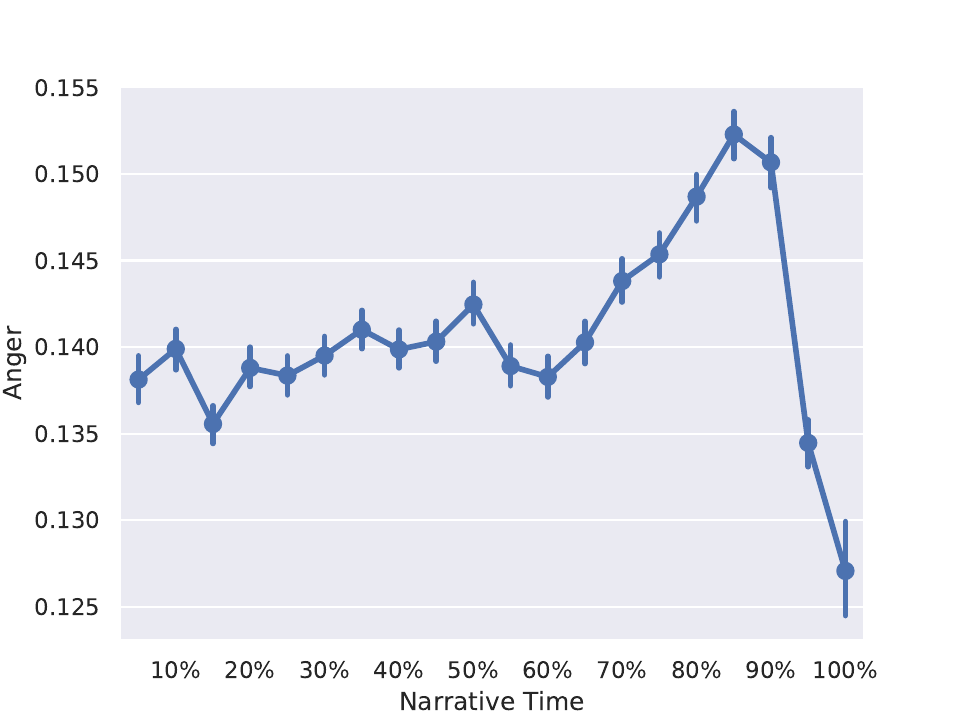}
        \caption{Anger}
        \label{fig:anger_narr}        
    \end{subfigure}
    \caption{\textit{Emotionality} increases, but specific emotions show non-linear trajectories over narrative time (95\% bootstrap confidence interval).}
    \label{fig:emotions_narr}
\end{figure}

We find that the emotional trajectories of performances are, in fact, structured over narrative time.
Figure~\ref{fig:neutral_decline} shows that emotionality increases over narrative time. We examine also the trajectory of \textit{specific} emotions across films (figs.~\ref{fig:joy_narr},\ref{fig:sadness_narr},\ref{fig:anger_narr}). We find that joyful performances follow a U-shaped curve, with a steep increase towards the end, as movies resolve. Like \citet{hipsonEmotionDynamicsMovie2021}, we find that negative-valence emotions like sadness and anger decrease at the end. Further, anger peaks 85\% into the film, reminiscent of a climax-resolution structure. %

\subsection{Evolving emotionality}
Subscribing to a particular categorization of emotions can be restrictive; in the remainder of the paper, we explore emotional performance, but depart from analyzing specific emotional labels. First, we study performance at a less granular level, focusing on the concept of \textit{emotionality} as the proportion of utterances with \textit{any} emotion label.\footnote{The model achieves an F1 score of 0.69 on the neutral label.} We measure how emotionality has changed historically over the decades spanned by our corpus.
Emotional shifts have been identified in English fiction books: \citet{morinBirthCoolTwocenturies2017} find that the content of those stories have experienced a decline in emotional expression.
Within cinema, David Bordwell has written about how shorter shot lengths and tighter framing serve to intensify the \textit{visual} style in more recent films compared to earlier ones \cite{bordwellIntensifiedContinuityVisual2002}. We ask whether there is a similar shift in performance: is there an intensification of emotion that matches the visual intensification of film, or perhaps an emotional cooling in performance that matches the findings in English fiction?

\begin{figure}
    \centering
    \includegraphics[width=0.7\linewidth]{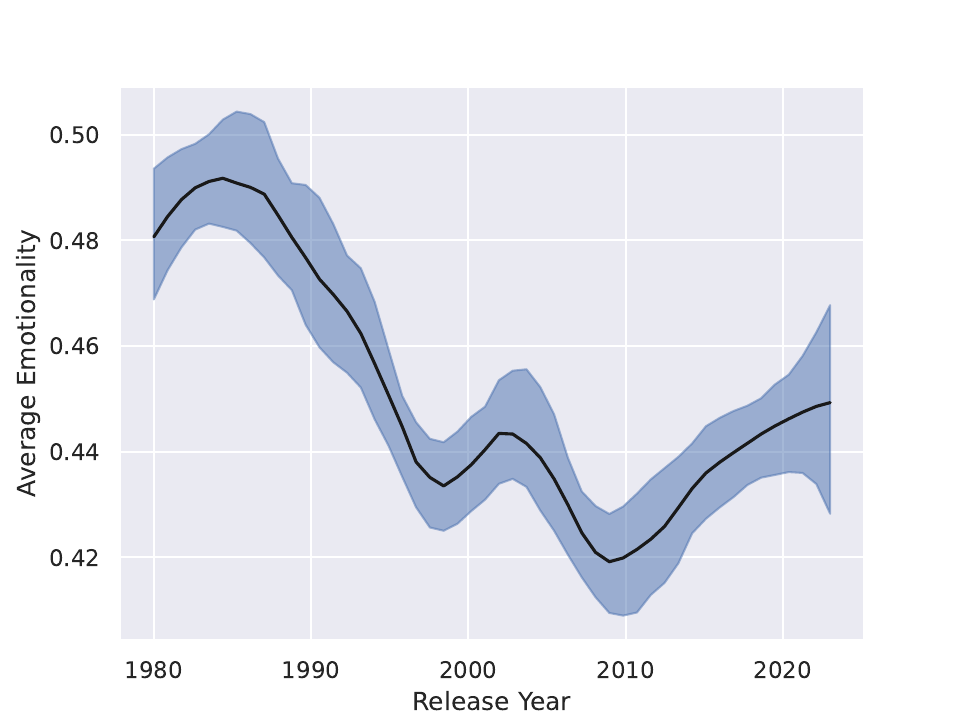}
    \caption{Emotionality is higher in older films (95\% bootstrap confidence interval).}
    \label{fig:neutral_increase}
\end{figure}

When we split the data by release year, we find a mild effect that earlier films have a higher proportion of emotional utterances compared to later ones, with emotionality hitting a minimum around 2010 (see Fig.~\ref{fig:neutral_increase}). However, the question remains whether the emotional \textit{content} is changing (as Morin and Acerbi find in literature) or if the \textit{style} with which words are being uttered is changing.

To disentangle the effects of shifting content and shifting style, we consider the change in emotionality over the years within the \textit{semantically equivalent} phrase groups. If it is indeed the writing, and not the performance, that drives this shift in emotionality, we should see little change within a phrase group. However, when we look at the 511 phrases that are used in all 43 years of the dataset, we find that a fixed-effects regression shows a slightly negative, statistically significant correlation between the year and emotionality even \textit{within} phrase groups ($R^2=0.048, F(1,21461), p < 0.001$).

Though this result is seemingly at odds with Bordwell's finding that visual style intensifies, it is also possible that they are harmonious. In Hollywood film, the close-up shot has always been associated with emotional expression \cite{pendleburyCuttingCenturyInvestigation2014}. \citet{panovskyStyleMediumMoving1959} writes that close-ups provide a rich ``field of action'' that affords nuanced acting performances. These visual performances, which are almost imperceptible if viewed from a natural distance, provide an alternative to the spoken word as a channel of expression. Comparing to the stage, Panovsky writes the spoken word makes a stronger impression ``if we are not permitted to count the hairs in Romeo's mustache.''
As cinema further grows into its medium, Bordwell finds that close-up shots have indeed grown tighter on the subject. With an increase in the capacity for more nuanced performance in the visual channel, the emotionality of the spoken word need not bear so strong a burden.

\subsection{Measuring emotional range}

\textit{Range} is often said to be the mark of a great actor. Naremore writes about the importance of an actor splitting their character ``visibly into different aspects'', showing off emotional range \cite{naremoreActingCinema1988}. Kuleshov similarly stressed the actors must be able to create a full range of gestures to create complex meaning \cite{kuleshovKuleshovFilmWritings1974}. \citet{wilsonLevelsAchievementActing1951} takes this a step further and argues that the hallmark of great acting is projecting a character into complex situations.
In this section, we explore the \textit{limits} of range through the constraints that genre and script impose on emotional performance.

For this analysis, we construct a general measure of emotional range across a set of utterances $u_{1\ldots n}$. We characterize each utterance $u_i$ with a performance vector $\vec{v}_i$, which is a distribution over emotions, given by the predicted probability distribution from the speech emotion recognition model. This allows us to take a more nuanced view of performances as a mixture of emotions. We model the distribution from which the vectors $\vec{v}_{1\ldots n}$ are drawn as a Dirichlet, and find the parameters which maximize the likelihood of the observed vectors. We define emotional range as the entropy of this distribution: a higher entropy means there is greater variance in the distribution of performances, and a lower entropy signals lower emotional range.

One criticism of the Ekman emotional model lies in its construct validity: seven discrete emotion labels may be insufficient to characterize the space of emotions. Ideally, we would model a continuous space of ``performance''. 
In our previous analyses, we use these emotion labels as an intermediate between that ideal on one end, and sentiment analysis on the other.
Here, our measure of emotional range is agnostic to the meaning of specific emotion labels, and serves to demonstrate how emotion classification can be a useful proxy task through which we can analyze performance in a more continuous space.

\paragraph{Thrillers have the least range; family-friendly films have the most.}

\begin{figure}
    \centering
    \includegraphics[width=0.7\linewidth]{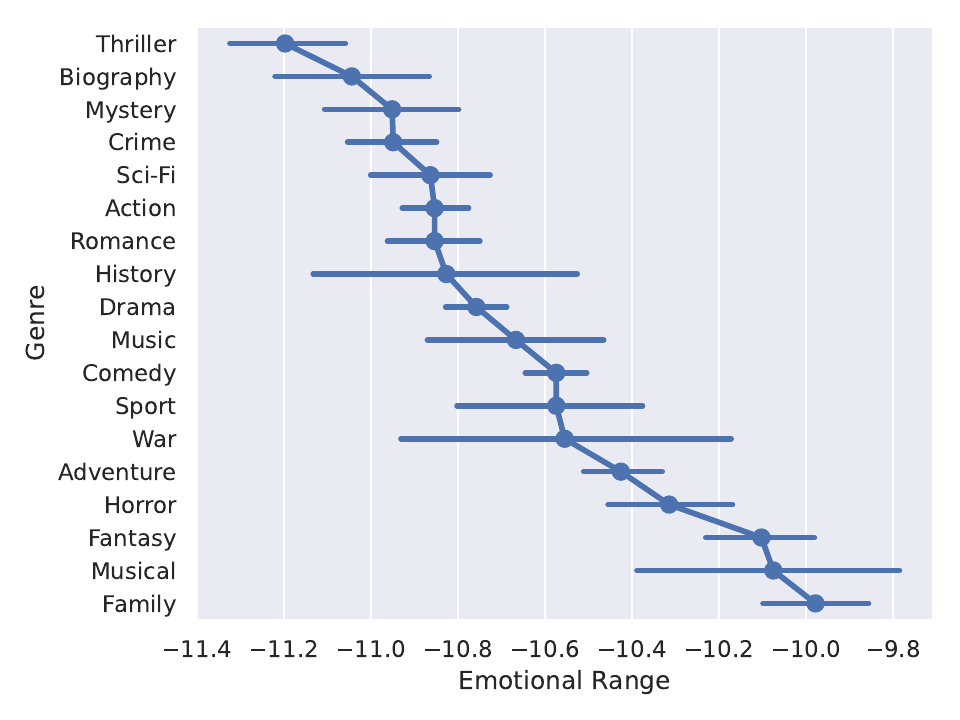}
    \caption{Relative emotional range for different film genres (95\% bootstrap confidence intervals).}
    \label{fig:genre_entropy}
\end{figure}

\citet{wilsonLevelsAchievementActing1951} speculates that some genres, like some types of comedy, have less capacity for emotional range than others. Previous work has shown that emotional \textit{arcs} are correlated with genre \cite{samothrakisEmotionalSentenceAnnotation2015}. We ask whether different genres are associated with different \textit{capacities} for emotional performance.

We calculate the emotional range for each movie, and find the average score for each genre. Genre information comes from IMDB, and we exclude genres with fewer than 30 films in our dataset.\footnote{Three genres were excluded: Western (19 films), Documentary (3), and Animation (2)} Figure~\ref{fig:genre_entropy} shows the average entropy across genres. Thrillers, biographies, and mysteries have the least emotional range; fantasy, musicals and family films rank highest. While it is difficult to attribute these results to a particular property of specific genres, these findings show that some genres have more constrained or consistent emotional registers than others.

\paragraph{Functional phrases have less capacity for emotional range.}
\citet{naremoreActingCinema1988} references Goffman when theorizing about performance: actors draw on and play against the interactional norms with which we as audience are already familiar. We ask if this bears out in our data. Does the emotional range of dialogue phrases reflect their discursive properties?

To study this, we measure the emotional range in dialogue. Because we tie specific performances to the words that are spoken, we can identify instances across the corpus when a given phrase was uttered. We isolate the 2,656 phrase groups that are uttered at least 50 times across our dataset. For each phrase group, we calculate the emotional range of its utterances.

\begin{table}[h]
  \centering
  \renewcommand{\arraystretch}{1.2} %
  \caption{Dialogue phrase groups with the highest and lowest emotional range scores. The table shows a representative phrase from each group.}
  \begin{tabular}{p{5cm}cp{5cm}c}
    \toprule
    \multicolumn{2}{c}{\textbf{Low Emotional Range}} & \multicolumn{2}{c}{\textbf{High Emotional Range}} \\ 
    \cmidrule(lr){1-2} \cmidrule(lr){3-4}
    \textbf{Phrase} & \textbf{Entropy} & \textbf{Phrase} & \textbf{Entropy} \\ 
    \midrule
    ``Could I ask you something?''& -17.02 & ``All rise.''& -7.85 \\
    ``This is your captain speaking.''& -16.56 & ``Are you out of your mind?'' & -7.88 \\
    ``Is that okay?''& -16.53 & ``What the fuck wrong with you?'' & -7.99 \\
    ``Can I get something for you?''& -16.17 & ``You're alive.'' & -8.32 \\
    ``Can I get something to drink?''& -16.16 & ``You saved my life.'' & -8.34 \\
    ``Hey, what can I get you?''& -15.91 & ``Don't you understand?'' & -8.34 \\
    ``You wanna come?''& -15.65 & ``Don't be so afraid.'' & -8.39 \\
    ``Yeah, that's good.''& -15.36 & ``You son of a bitch.'' & -8.40 \\
    ``Any questions?''& -15.26 & ``You scared the shit out of me!''& -8.44 \\
    ``That's correct.''& -15.25 & ``Ow.''& -8.44 \\
    \bottomrule 
  \end{tabular}
  \label{tab:phrase_entropies}
\end{table}

Table~\ref{tab:phrase_entropies} shows phrases with the highest and lowest entropies. By inspecting the phrases at either end of the spectrum, we find qualitative differences in the kinds of phrases that have higher and lower emotional range: the capacity for emotional variance reflects the discursive flexibility of the words being spoken.
Phrases with low range are functional and generally part of highly directed interactions: most phrases are either yes-or-no questions or answers to them. Phrases with high emotional range, on the other hand, mostly have more open-ended, evaluative discursive functions. In these cases, the prosody or intonation of speech can easily lend color to the statement being made. ``You're alive'' can be said with joy or relief to a loved one, as Marty McFly to his mentor Doc in \textit{Back to the Future} (1985), or with anger at the sight of an enemy, as Lord Norinaga greets Walker in \textit{Teenage Mutant Ninja Turtles III} (1993).

\section{Discussion and limitations}
\label{sec:limitations}

With this work, we demonstrate that films can, and should, be studied as performance texts. We tie our findings to both film theory and other computational work on narratives. Here, we discuss some limitations of the current study.

\paragraph{Measuring emotions.} We follow a vast body of previous work within natural language processing \cite{zhaoM3EDMultimodalMultiscene2022,zahiriEmotionDetectionTV2017}, affective computing \cite{bussoIEMOCAPInteractiveEmotional2008a,caoCREMADCrowdsourcedEmotional2014} and computational literary studies \cite{samothrakisEmotionalSentenceAnnotation2015,barrosAutomaticClassificationLiterature2013} in using Ekman's basic emotions. However, the validity of this model has been questioned \cite{plaza-del-arcoEmotionAnalysisNLP2024}.

First, there are doubts about the ecological validity of emotion recognition, especially as most speech emotion recognition datasets contain acted emotion as opposed to natural emotion. We note that, unlike much affective computing work, we use emotion recognition models trained on acted speech to make inference on acted speech. The professionally-produced, acted speech in the MELD dataset is well-suited to our data, which is also professionally-produced and acted. Indeed, we find that  performance is similar between MELD and our in-domain evaluation data.

Another criticism lies in the cultural relativity of emotion. Though Ekman argues that the basic emotions are universal, he acknowledges there may be cultural differences in the emotions elicited in a given context.
It is reasonable to suppose that viewers' normative knowledge also influences the interpretation of these emotions. We focus on contemporary American film in both our analysis and training data, holding at least the intended cultural audience constant. Cultural variation in performance is a ripe area for future work, as cultural differences exist in not only the production and interpretation of emotion, but also in theories of acting.

Aside from these specific criticisms of the Ekman model, the low interannotator agreement in both our evaluation set as well as other datasets, including MELD, suggest that this model for emotion may remain too coarse to precisely describe the data. Work in both affective psychology and NLP have attempted to address this by using more fine-grained classes \cite{demszkyGoEmotionsDatasetFineGrained2020, cowenMappingPassionsHighDimensional2019} or a continuous spaces of emotion \cite{russellCircumplexModelAffect1980,cowenMappingPassionsHighDimensional2019}. While we used the Ekman model due to the availability of training data as well as to provide comparison with previous studies of emotion narratives, alternative emotion models may prove useful in future work.

\paragraph{A question of authorship.}

In cinema, the performance that audiences see on screen is co-created by the actor, the director, and the editor. \citet{baronReframingScreenPerformance2008} describe the conventional wisdom within film analysis to be that cinematic performances are made in the cutting room. ``True'' acting happens on the stage. Though our work studies film as performance text, it does not disentangle the processes through which the performance is constructed. It is about the performance as viewed, but not about the choices made by actors as separate from the director or editor. Our work makes the point that performance carries meaning worth studying, and opens the door for future computational work that explores its authorial roots.

\paragraph{Embodied erformance.}

Finally, we examine only performance as enacted through speech. This is perhaps the modality that lies closest to the script, and allows us to apply a variationist approach to studying the relationship between performance and text, but of course performance includes not just speech but also gesture, posture, facial expression, and more. Visual description has been found to be more useful for aligning narrative events than dialogue \cite{zhouEvaluationAlignmentMovie}, and quantitative analysis of theater performance has found narratively meaningful patterns in movement \cite{escobarvarelaQuantitativeCloseAnalysis2017}. Film is a multimodal medium that deserves analysis in all its modalities. We hope that our work examining film across the speech and text can serve as a basis for more work that examines performance as embodied visually.

\section{Conclusion}

In this paper, we explore the relation between film as narrative text and as performance text. Using a novel parallel dataset of speech and text from popular contemporary American film, we develop computational methods to measure how emotional prevalence and emotional range vary by both textual factors of narrative time and dialogue, as well as contextual factors of release year and genre. We hope this work inspires further multimodal studies of \textit{performance} in computational film analysis.

\begin{acknowledgments}

The research reported in this article was supported by funding from Mellon Foundation and the National Science Foundation (IIS-1942591 and DGE-2146752). We thank Jacob Lusk and Lucy Li for insightful discussion and feedback.

\end{acknowledgments}

\bibliography{bibliography}

\appendix

\end{document}